\pdfoutput=1

\documentclass[11pt]{article}

\usepackage{ACL2023}

\usepackage{times}
\usepackage{latexsym}

\usepackage[T1]{fontenc}

\usepackage[utf8]{inputenc}

\usepackage{microtype}

\usepackage{graphicx}
\usepackage{booktabs}
\usepackage{multicol}
\usepackage{multirow}
\usepackage{amsmath}
\usepackage{amssymb}
\usepackage{bbding}
\usepackage{subfigure}
\usepackage{colortbl}
\usepackage{pifont}
\usepackage{bbm}
\usepackage{makecell}
\usepackage{bbding}
\usepackage{mathrsfs}
\usepackage{bm}
\usepackage{tabu}
\usepackage{dsfont}
\usepackage{tcolorbox}
\usepackage{xcolor}
\usepackage{soul}
\usepackage{framed}
\definecolor{shadecolor}{rgb}{0.92,0.92,0.92}
\usepackage{arydshln}
\usepackage{textcomp}
\newcommand{\textpermyriad}{\text{\textperthousand}\kern-0.25em\text{\textperthousand}}

\newtcolorbox{mybox}[1]{colback=blue!5!white,
colframe=blue!80!black,fonttitle=\bfseries,
title={#1}}

\newcommand{\ourmethod}{\textsc{FuzzCoder}}
\newcommand{\benchmark}{Fuzz-Bench}
\newcommand{\instruct}{Fuzz-Instruct}

\title{\ourmethod{}: Byte-level Fuzzing Test via Large Language Model}

\author{
  \fontsize{11pt}{0pt}\selectfont Liqun Yang\textsuperscript{\rm 1}, 
  Jian Yang\textsuperscript{\rm 1}\thanks{\ \ Corresponding author.},
  Chaoren Wei\textsuperscript{\rm 1}, 
  Guanglin Niu\textsuperscript{\rm 2},
  Ge Zhang\textsuperscript{\rm 3,5},
  Yunli Wang\textsuperscript{\rm 1},   \\
   \fontsize{11pt}{0pt}\selectfont 
  \textbf{Linzheng Chai}\textsuperscript{\rm 1}, 
  \textbf{Wanxu Xia}\textsuperscript{\rm 1},
  \textbf{Hongcheng Guo}\textsuperscript{\rm 1},
  \textbf{Shun Zhang}\textsuperscript{\rm 1},
  \textbf{Jiaheng Liu}\textsuperscript{\rm 1},
  \textbf{Yuwei Yin}\textsuperscript{\rm 1}, \\
  \fontsize{11pt}{0pt}\selectfont 
  \textbf{Junran Peng}\textsuperscript{\rm 4},
  \textbf{Jiaxin Ma}\textsuperscript{\rm 6},
  \textbf{Liang Sun}\textsuperscript{\rm 1} 
  \textbf{Zhoujun Li}\textsuperscript{\rm 1} \\
  \textsuperscript{\rm 1}Beihang University; \textsuperscript{\rm 2}University of British Columbia; \textsuperscript{\rm 3}University of Waterloo\\
    \textsuperscript{\rm 4}University of Science and Technology Beijing; \textsuperscript{\rm 5}M-A-P;\\
  \textsuperscript{\rm 6}Beijing University of Posts and Telecommunications\\
  weichaoren@buaa.edu.cn; 
}

\begin{document}

\maketitle

\begin{abstract}
Fuzzing is an important dynamic program analysis technique designed for finding vulnerabilities in complex software. Fuzzing involves presenting a target program with crafted malicious input to cause crashes, buffer overflows, memory errors, and exceptions. Crafting malicious inputs in an efficient manner is a difficult open problem and the best approaches often apply uniform random mutations to pre-existing valid inputs. 
In this work, we propose to adopt fine-tuned large language models (\ourmethod{}) to learn patterns in the input files from successful attacks to guide future fuzzing explorations. 
Specifically, we develop a framework to leverage the code LLMs to guide the mutation process of inputs in fuzzing. The mutation process is formulated as the sequence-to-sequence modeling, where LLM receives a sequence of bytes and then outputs the mutated byte sequence. \ourmethod{} is fine-tuned on the created instruction dataset (\instruct{}), where the successful fuzzing history is collected from the heuristic fuzzing tool. \ourmethod{} can predict mutation locations and strategies locations in input files to trigger abnormal behaviors of the program. Experimental results show that \ourmethod{} based on AFL (American Fuzzy Lop) gain significant improvements in terms of effective proportion of mutation (EPM) and number of crashes (NC) for various input formats including ELF, JPG, MP3, and XML.\footnote{\url{https://github.com/weimo3221/FUZZ-CODER}}
\end{abstract}

\section{Introduction}
Fuzzing test \cite{guo2018dlfuzz,xie2022docter,wei2022free,cummins2018compiler,manes2019art,li2018fuzzing}, a dynamic software testing technique, has emerged as a powerful method for uncovering vulnerabilities and defects within software applications. Fuzzing frameworks like AFL (American Fuzzy Lop) and libFuzzer have become industry standards, while researchers further explore advanced strategies like evolutionary fuzzing and hybrid approaches to enhance test case generation and code coverage. As the intricacy of software systems escalates, fuzzing continues to evolve, proving its essential role in the realm of software development and security testing.

\begin{figure}[t!]
\begin{center} \includegraphics[width=1.0\columnwidth]{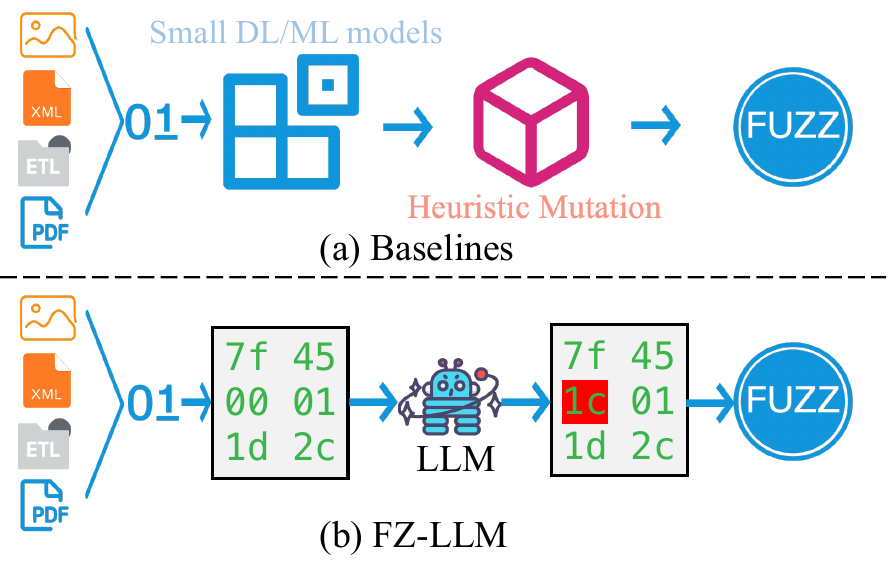}
	\caption{Comparison between the standard byte-level fuzz test and our proposed method.}
	\label{fig:intro}
        \vspace{-10pt}
\end{center}
\end{figure}

Based on neural network architectures like RNNs and GANs~\cite{goodfellow2016deep_learning}, this line of research has shown potential in improving test case generation, increasing code coverage, and detecting elusive vulnerabilities. Trained on billions of lines of code, large language models (LLMs) have shown exceptional aptitude in various software engineering tasks in code generation \cite{codellama,qwen,deepseek_coder}, program repair \cite{program_repair_review,guo2023owl}, and fuzzing \cite{fuzz4all,zero_shot_fuzzer,llm_fuzz_survey,yang2024seq2seq}. The rigorous pre-training on vast code datasets forms the cornerstone of the capabilities of LLM in code generation and comprehension, even for the encoded byte sequence. Byte level byte pair encoding (BBPE) tokenizer \cite{bbpe,bGPT,gpt2} have become the standard practices for state-of-the-art LLMs, which brings powerful understanding and generation capability for byte-like data. Moreover, these LLMs can be further optimized through fine-tuning or prompting to enhance their proficiency in specific domains. 
\textit{However, how to effectively leverage instruction fine-tuning (IFT) to inspire LLMs to help byte-based mutation for the fuzzing test still requires further exploration.}

In this paper, we investigate the feasibility of leveraging code LLM to develop a framework, guiding the mutation process of inputs in fuzzing. The mutation process is formulated as the sequence-to-sequence modeling, where LLM receives a byte sequence and then outputs the mutated byte sequence. The LLM is fine-tuned on the created instruction dataset, where the successful fuzzing history is collected from the heuristic fuzzing tool. In Figure \ref{fig:intro}, the instruction corpus is coupled into pairs comprised of original inputs and successfully mutated inputs. \ourmethod{} aims at identifying the most possible bytes within input files for mutations. To gather the instruction dataset \instruct{}, we initially adopt standard fuzzing methods to record mutation instances that yield new code coverage or trigger crashes. \instruct{} then serves to train \ourmethod{} based on different code foundation models to guide towards generating promising mutated inputs. While our methodology is adaptable to various fuzzing frameworks, we apply it specifically to the state-of-the-art AFL, which introduces random mutations into a batch of seed input files and curates a queue of new inputs, which are effective in tracing new code executions.

Our proposed method is evaluated on the benchmark \benchmark{}, comprised of 8 programs: \texttt{NM\_ELF}, \texttt{READ\_ELF}, \texttt{OBJDUMP\_ELF}, \texttt{LINT\_XML}, \texttt{MP3GAIN\_MP3}, \texttt{IMAGEMAGICK\_GIF}, \texttt{SPLIT\_TIFF}, and \texttt{TRAN\_JPEG}. \benchmark{} accepts the different format inputs, including ELF, XML, MP3, and GIF. \ourmethod{} significantly improves line coverage and branch coverage compared to the previous strong baselines. Further, we observe that \ourmethod{} triggers more new paths or the frequency of code blocks found during fuzz testing due to the effective mutation prediction of the understanding capability of the code LLM.

The key contributions are summarized as:
\begin{itemize}
\item We formulate the fuzzing test as a sequence-to-sequence paradigm and then introduce the generation model to attack vulnerable positions by selecting proper mutation positions and strategies. The data in any format is first converted into a sequence of bytes as the input of LLMs. Then, the code LLM will decide the possible mutation strategies and positions.  
\item We construct a complete framework to fine-tune the code LLMs with the help of the collected instruction corpora \instruct{}. To effectively evaluate the performance of different models, we construct a fuzzing test benchmark \benchmark{} comprised of 8 programs, which accept different formats of data (e.g. ELF, JPG, MP3, and XML).
\item The experimental results on created benchmark \benchmark{} (simulation using AFL) demonstrate the fine-tuned \ourmethod{} significantly improves the effective proportion of mutation (EPM) and triggers more program crashes compared to the previous baselines.

\end{itemize}

\section{Preliminary: Fuzzing Test}
Fuzzing is a robust software testing technique designed to uncover vulnerabilities and flaws in computer programs, primarily by subjecting them to a barrage of unexpected and often invalid inputs. The fuzzing test can be mathematically represented as follows:
\begin{BigEquation}
\begin{align}
    \mathcal{F}(T, g(x)) = R
    \label{problem_definition}
\end{align}
\end{BigEquation}where $\mathcal{F}(\cdot, \cdot)$ represents the fuzzing process receiving mutation of input test cases. $T$ is the target software or program subjected to the fuzzing test.
$I$ represents the input test cases, which are typically malformed, unexpected, or random data. $g(x)$ is the mutation format of the original input $x$. $R$ stands for the results or observations obtained during the fuzzing test, which may include system crashes, error messages, or other unexpected behaviors in the target software.

American Fuzzy Lop\footnote{\url{https://github.com/google/AFL}} (AFL) is a widely used automated vulnerability mining tool, which finds security vulnerabilities in software programs through fuzzy testing techniques. Fuzzy testing is a black-box testing methodology that injects random or semi-random data into program inputs to detect anomalous behavior and potential vulnerabilities in the program. In AFL, \textbf{mutation} refers to the generation of new fuzzy test inputs by modifying the input samples, which is a core component of AFL fuzzy testing. Its mutation strategy employs a range of random and semi-randomized mutation techniques to create a diversity of test inputs. Let $x^{(i)} \in \{x^{(1)},\dots,x^{(n)}\}$ denote the seed test input from the initial pool comprised of $n$ test cases, we leverage the NLP techniques to generate the mutated test case $z^{(i)}$. Different from the rule-based mutation, we use a generation model to obtain variant samples for fuzzy testing by predicting variant locations and variant types. Specifically, $x^{(i)}=\{x^{(i)}_{1},\dots,x^{(i)}_{m}\}$ is $m$ bytes input sequence, the prediction model $\mathcal{M}$ chooses $k$ mutation positions $p=\{p_1,\dots,p_{k}\}$ and their corresponding mutation strategies $s=\{s_1,\dots,s_{k}\}$ to modify the original test case $x^{k}$ into $z^{k}$. The process can be described as:
\begin{BigEquation}
\begin{align}
    P(p, s|x^{(i)}) = \prod_{j=1}^{m}P(p_j,s_j|x^{(i)},p_{< j},s_{< j};\Theta)
\end{align}
\end{BigEquation}where $p_{<j}=(p_1,\dots,p_{j-1})$ and $s_{<j}=(s_1,\dots,s_{j-1})$. $p_j$ and $s_j$ represent the $j$-th mutation position and mutation strategy respectively predicted by the previous context $p_{<j}$ and $s_{<j}$ sequentially and the original test case $x^{(i)}$.

\begin{figure*}[t!]
\begin{center} \includegraphics[width=0.7\textwidth]{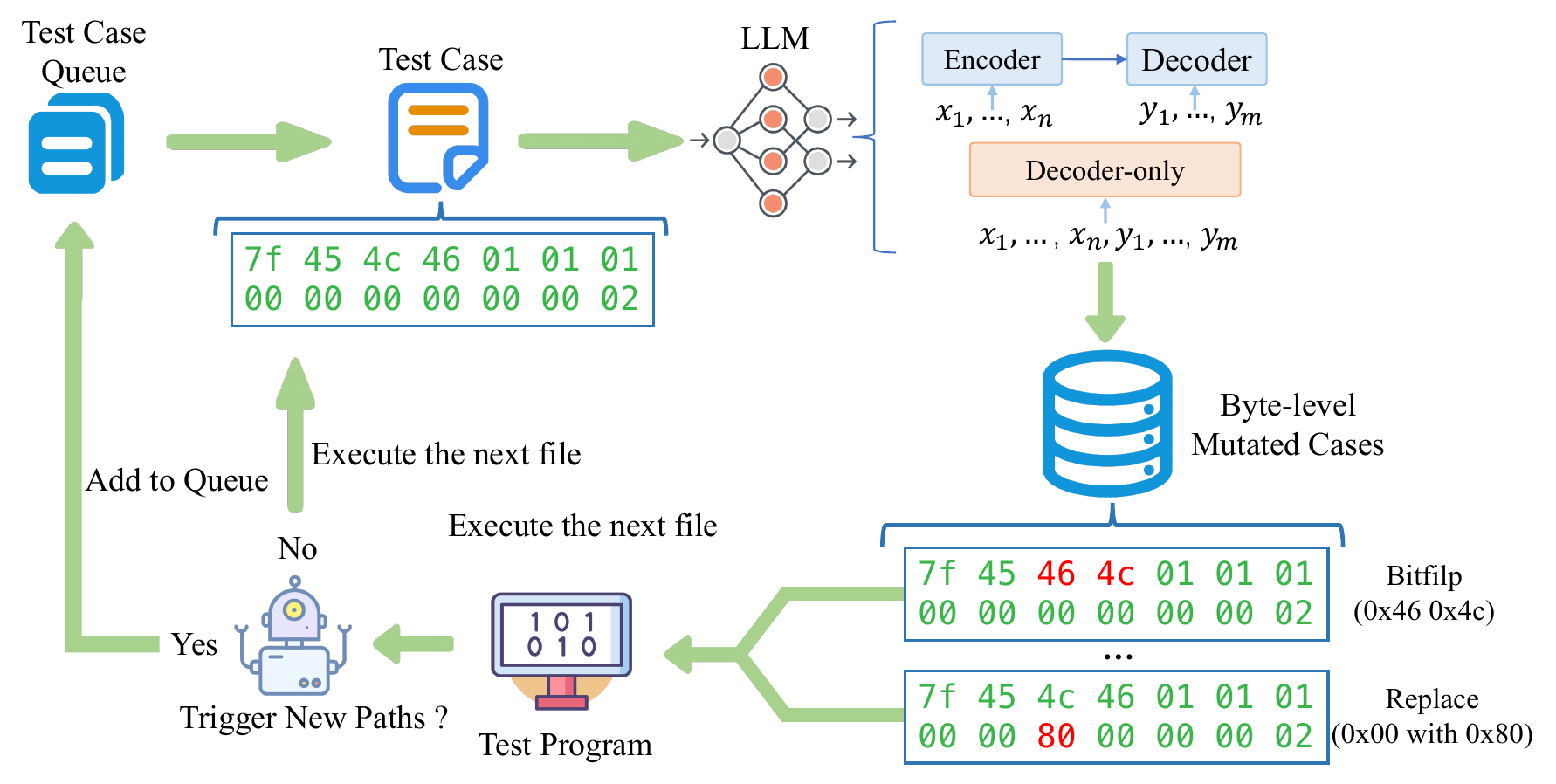}
	\caption{The workflow of the fuzzing test with fine-tuned LLMs \ourmethod{}.}
	\label{framework}
\vspace{-15pt}
\end{center}
\end{figure*}

\section{\benchmark{}}

We introduce 8 fuzzing datasets: \texttt{NM\_ELF}, \texttt{READ\_ELF}, \texttt{OBJDUMP\_ELF}, \texttt{LINT\_XML}, \texttt{MP3GAIN\_MP3}, \texttt{IMAGEMAGICK\_GIF}, \texttt{SPLIT\_TIFF}, and \texttt{TRAN\_JPEG}, which accept the different format inputs, including the ELF, XML, MP3, and GIF format. The program subjected to the fuzzing test originates from the FuzzBench\footnote{\url{https://github.com/google/FuzzBench}} and previous works\footnote{\url{https://github.com/fdu-sec/NestFuzz}}.

Here, we describe the details of each dataset. For \texttt{LINT\_XML}, the program parses one or more XML files and prints various types of output, depending upon the options selected. It is useful for detecting errors both in XML code and in the XML parser itself. For \texttt{READ\_ELF}, the program reads and displays information about the contents of ELF (executable and linkable Format) format files, which include executables, target files, and shared libraries. For \texttt{NM\_ELF}, the program displays symbol table information in target files (including executables, target files, and shared libraries). The symbol table contains symbols defined and referenced in the program (e.g., variable names, function names, etc.) and their associated attributes. For \texttt{OBJDUMP\_ELF}, the program displays various information from object files (including executable files, target files, and shared libraries), such as disassembled code and section table information. For \texttt{MP3GAIN\_MP3}, the program adjusts the volume of MP3 audio files, which aims to balance and normalize the volume of MP3 files so that they sound more consistent when played without noticeable volume differences. For \texttt{IMAGEMAGICK\_GIF}, the program is a tool in ImageMagick for processing various image files (including JPG, PNG, GIF, etc.). It can get information about the image, adjust the image, and process it. For \texttt{SPLIT\_TIFF}, it splits a TIFF file containing multiple images into multiple separate TIFF files, each file containing a frame or page from the input file. For \texttt{TRAN\_JPEG}, it can rotate JPG images 90 degrees, 180 degrees or 270 degrees clockwise. JPG images can also be cropped, optimized, etc.

\paragraph{Data Construction}
For different programs, we need to collect the data used for LLMs separately by fuzzing the programs with heuristic methods, where the baseline is denoted as AFL. Through the simulation of the original AFL, we can collect the $k$ valid mutations $\{(p_1,s_1),\dots,(p_k,s_k)\}$ for the specific test case $x$. Then, we can construct the supervised training pair $(x,p,s)$ comprised of the input test case $x$, valid mutation positions $p$, and the corresponding strategies $s$. For each dataset, we can obtain the corresponding instruction corpus $D_{t}=\{I^{(i)}, x^{(i)},y^{(i)}\}_{i=1}^{N_{t}}$ ($1 \le t \le T=8$, $T$ is the number of the programs, $N_{t}$ is the training data size of the program $t$, and $I^{(i)}$ is the instruction) and merge them as the whole dataset $D=\{D_{t}\}_{t=1}^{T}$.

Given the specific test case, there exist different valid mutation strategies to successfully fuzz the program (e.g. the mutation leads to the program crash or triggers a new execution path). We can gather the valid mutation pairs together as the target sequence. i.e., valid $(p_{i}, s_{i})$ pairs of the test case. In the following example, if its valid $(p_{i}, s_{i})$ pairs are $(1, 2)$ and $(1, 3)$, it denotes that the $2$-th and $3$-th token in the hexadecimal sequence will perform $1$-th operation to cause crash of the program.
the final expression can be described as follows:
\begin{mybox}{Data Collection}
    Byte Input: 0x3c\ 0x21\ 0x44\ 0x4f\ 0x43\\
    Mutation\ strategies: [$(1, 2)$, $(1,3)$]
\end{mybox}

The queue of input sequences $Q$ is used to store input test cases (test cases). When the fuzzing process (e.g. AFL) starts, it automatically selects and mutates input data based on the response of the target program to better explore potential program paths and boundary conditions. $Q$ contains input files that successfully caused the program to execute different paths during testing. These input files are considered valid because they cause program execution to enter new code paths or trigger specific error conditions. To collect as much mutation data as possible for each program, each program is fuzzed multiple times.

\paragraph{Data Split}
Since the training of the model requires a training set and a valid set, we randomly select 90\% of the samples as the training set and 10\% of the data as the valid set. The number of samples is described as:
\begin{table}[h!]
\centering
\resizebox{1\columnwidth}{!}{
\begin{tabular}{l|ccccc}
    \toprule
    Benchmark                   & Train & Test & Program            & Input    & Option\\  \midrule
    \texttt{NM\_ELF}            & 4534 & 504   & \textit{nm-new}    & ELF            & -a @@\\ 
    \texttt{READ\_ELF}          & 4167 & 464   & \textit{readelf}   & ELF            & -a @@\\
    \texttt{OBJDUMP\_ELF}       & 4009 & 446   & \textit{objdump}   & ELF            & -x -a -d @@\\ 
    \texttt{LINT\_XML}          & 5442 & 605   & \textit{xmllint}   & XML            &  --valid --recover @@\\ 
    \texttt{MP3GAIN\_MP3}       & 1431 & 150   & \textit{mp3gain}   & MP3            & @@\\
    \texttt{IMAGEMAGICK\_GIT}   & 6477 & 720   & \textit{magick}    & GIF            & identify @@\\
    \texttt{SPLIT\_TIFF}        & 4136 & 459   & \textit{tiffsplit} & TIFF           & @@\\  
    \texttt{TRAN\_JPEG}         & 1376 & 153   & \textit{jpegtran}  & JPEG           & @@\\
    \bottomrule
    \end{tabular}
}
\caption{Statistics of the different benchmarks.}
\vspace{-10pt}
\end{table}

\paragraph{Simulation Environment}
We incorporate the generation model into the AFL framework to support the fuzzing with LLM. The simulation environment is Ubuntu 18.04.6 LTS, Intel Xeon Processor (Skylake, IBRS), A100-PCIE-40GB, AFL-2.57b\footnote{\url{https://github.com/google/AFL}}.

\section{Fuzzing Test via Generation Model}

\subsection{Input Encoding}
Our framework consists of a fuzzer and a model that highlights useful locations in an input file. During runtime. the fuzzer queries the model for each seed file and focuses mutations on the highlighted locations. Given an open-ended input file, we first convert the input file into a sequence of bytes $x^{(i)}$ in Figure \ref{framework} (hexadecimal sequence). Then, the generation model should predict the mutation positions $p=\{p_1,\dots,p_k\}$ and the mutation strategies $s=\{s_1,\dots,s_k\}$, where the $s_{k}$ is the corresponding mutation strategy of the position $p_k$. To jointly model the mutation position and strategy, the prediction sequence $y=(y_1,\dots,y_{2k})$ can be described as:
\begin{BigEquation}
\begin{align}
       y = (p_1,s_1,\dots,p_k,s_k)
    \label{target_sequence}
\end{align}
\end{BigEquation}where the model first predicts the mutation position $p_k$ and then output the corresponding strategy $s_k$.

\subsection{Encoder-Decoder Framework}
Given the source inputs $D_{src}$ and target predictions $D_{trg}$, the encoder of the encoder-decoder-based \ourmethod{} first receives the original input $x$ and encodes it into the hidden states $H_{enc}$ with the bidirectional attention mechanism.
\begin{BigEquation}
\begin{align}
    H_{e} &=\mathcal{S}(x, \mathcal{M}_e) = \overset{A}{\underset{a=1}{\big\|}} \texttt{Softmax}\left(\frac{QK^T}{\sqrt{d_k}} \otimes \mathcal{M}_{e}\right) V 
    \label{encoder_decoder}
\end{align}
\end{BigEquation}where $A$ is the number of attention heads
Then, the decoder predicts the target tokens sequentially based on $H_{e}$. 

\subsection{Decoder-only Framework}

Given the source inputs $D_{src}$ and target predictions $D_{trg}$, the encoder of the encoder-decoder-based \ourmethod{} first receives the original input $x$ and encodes it into the hidden states $H_{enc}$ with the bidirectional attention mechanism.
\begin{BigEquation}
\begin{align}
       H_{d} &=\mathcal{S}(x, \mathcal{M}_d) = \overset{A}{\underset{a=1}{\big\|}} \texttt{Softmax}\left(\frac{QK^T}{\sqrt{d_k}} \otimes \mathcal{M}_{d}\right) V
\end{align}
\end{BigEquation}where $A$ is the number of attention heads
The decoder predicts the target tokens sequentially based on $H_{e}$ with the casual mask $\mathcal{M}_d$.

\subsection{Mutation Strategy Prediction}

For each mutation position $p_{j}$, we use the generation model to infer the possible mutation strategy for the position. 12 candidate mutation strategies are provided for each position, including:
(1) bitflip 1/1: perform bitfilp on a bit randomly.
(2) bitflip 2/1: perform bitfilp on two neighboring bits randomly.
(3) bitflip 4/1: perform bitfilp on four neighboring bits randomly.
(4) bitflip 8/8: randomly select a byte and XOR it with 0xff.
(5) bitflip 16/8: randomly select two neighboring bytes and XOR them with 0xff.
(6) bitflip 32/8: randomly select four neighboring bytes and XOR them with 0xff.
(7) arith 8/8: randomly select a byte and perform addition or subtraction on it (operands are 0x01~0x23).
(8) arith 16/8: randomly select two neighboring bytes and convert these two bytes into a decimal number. Select whether to swap the positions of these two bytes. Perform addition or subtraction on it (operands are 1~35). Finally, convert this number to 2 bytes and put it back to its original position.
(9) arith 32/8: randomly select four neighboring bytes. Select whether to swap the positions of these four bytes. Convert these four bytes into a decimal number. Perform addition or subtraction on it. Finally, convert this number to 4 bytes and put it back to its original position.
(10) interest 8/8: randomly select a byte and replace it with a random byte.
(11) interest 16/8: randomly select two neighboring bytes and replace them with two random bytes.
(12) interest 32/8: randomly select four neighboring bytes and replace them with four random bytes.
%



\subsection{Jointly Training}
Since the mutation strategies and positions $y = (p_1,s_1,\dots,p_k,s_k)$ are our prediction goals, the supervised fine-tuning objective of \ourmethod{} can be described as:
\begin{BigEquation}
\begin{align}
\begin{split}
\label{Easy_first_language_model}
    \mathcal{L}_{m} = -\mathbb{E}_{x^{(i)},p^{(i)},s^{(i)} \in D_{src}}\log P(p^{(i)},s^{(i)}|x^{(i)})
\end{split}
\end{align}
\end{BigEquation}where $x^{(i)}$ is the $i$-th original input from the collected dataset. $p = (p_1,\dots,p_k)$ is the predicted mutation positions and $s = (s_1,\dots,s_k)$ is the mutation strategies.

\subsection{Incorporating LLMs into Fuzzing Test}
\begin{figure}[t!]
\centering
\includegraphics[width=1.0\linewidth]{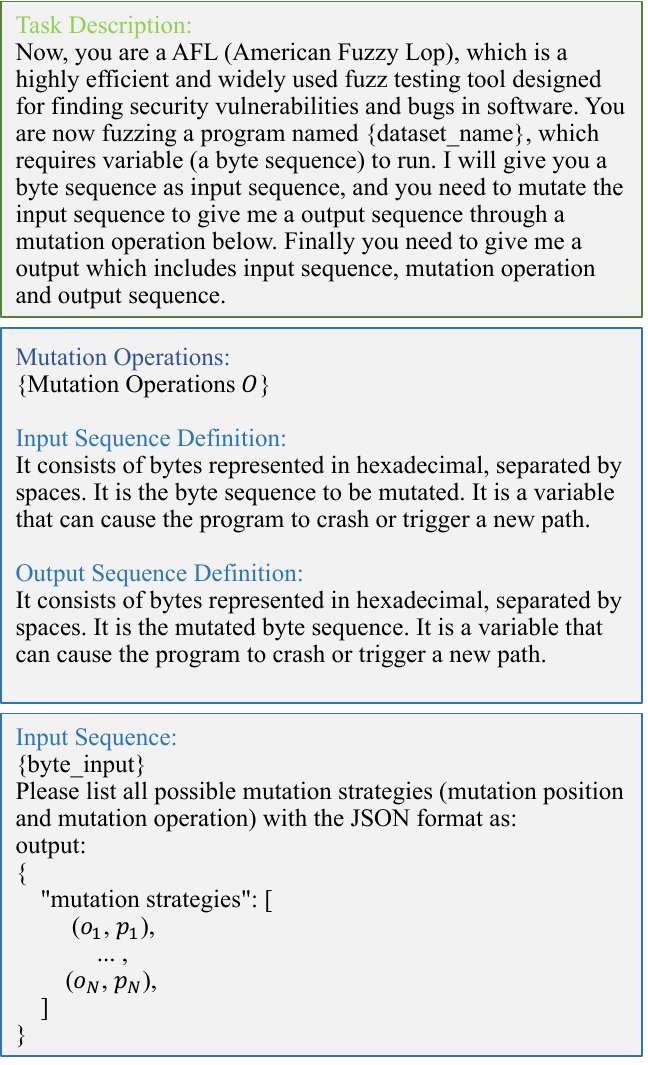}
\caption{The prompt to get mutation positions and strategies of \ourmethod{}. }
\vspace{-10pt}
\label{intro}
\end{figure}

\begin{table*}[t!]
\begin{center}
\resizebox{1.0\textwidth}{!}{
\begin{tabular}{llcccccccccccccc}
\toprule
Method  & Base & Size & bitflip 1/1   & bitflip 2/1 & bitflip 4/1   & bitflip 8/8   & bitflip 16/8 & bitflip 32/8 & arith 8/8 & arith 16/8 & arith 32/8   & interest 8/8 & interest 16/8 & interest 32/8 & Avg. \\ 
\midrule
\rowcolor[rgb]{0.85,0.92,0.83} \multicolumn{16}{c}{ \texttt{READ\_ELF}}   \\ 
\midrule
AFL (Original) & - & - & 1.50 & 0.66 & 0.25 & 0.33 & 0.09 & 0.24 & 0.30 & 0.00 & 0.00 & 0.48 & 0.06 & 0.03 & 0.33\\
AFL (LSTM) & - & - & 1.37 & 1.11 & 0.97 & 0.00 & 0.00 & 0.00 & 2.49 & 0.00 & 0.00 & 0.00 & 0.00 & 0.49 & 0.54 \\
AFL (Transformer) & - & - & 1.11 & 1.04 & 1.02 & 1.61 & 0.00 & 0.90 & 3.99 & 0.22 & 0.30 & 2.34 & 1.98 & 0.82 & 1.28\\
\arrayrulecolor{lightgray}\hdashline
\ourmethod{} & StarCoder-2    & 7B & 3.42 & 0.92 & 1.28 & 2.45 & 0.12 & 0.15 & 0.63 & 0.12 & 0.05 & 0.45 & 2.41 & 0.34 & 1.03 \\
\ourmethod{} & StarCoder-2    &15B & 4.21 & 2.38 & 1.43 & 2.95 & 0.24 & 0.21 & 1.25 & 0.45 & 0.38 & 0.57 & 1.38 & 0.45 & 1.32 \\
\ourmethod{} & CodeLlama      & 7B & 3.82 & 2.24 & 1.45 & 2.01 & 0.17 & 0.33 & 1.36 & 0.19 & 0.43 & 1.24 & 0.95 & 0.91 & 1.26 \\
\ourmethod{} & DeepSeek-Coder & 7B & 1.98 & 1.73 & 0.66 & 3.13 & 0.08 & 0.24 & 2.92 & 0.22 & 0.25 & 1.48 & 1.82 & 2.05 & 1.38\\
\ourmethod{} & CodeQwen       & 7B & 3.00 & 1.41 & 2.07 & 1.09 & 0.66 & 0.97 & 5.86 & 0.37 & 0.37 & 0.73 & 0.54 & 1.15 & \cellcolor{red!25} \bf 1.52 \\
\ourmethod{} & CodeShell      & 7B & 2.08 & 2.42 & 1.34 & 3.81 & 0.54 & 0.55 & 2.45 & 0.55 & 0.02 & 0.45 & 0.25 & 1.23 & 1.31
\\ \arrayrulecolor{black}\midrule
\rowcolor[rgb]{0.85,0.92,0.83}\multicolumn{16}{c}{\texttt{OBJ\_DUMP}} \\ \midrule
AFL (Original)  & - & - & 2.07 & 0.89 & 0.43 & 0.43 & 1.35 & 1.93 & 0.31 & 0.08 & 0.01 & 0.79 & 0.21 & 0.11 & 0.72 \\
AFL (LSTM) & - & - & 1.26 & 4.20 & 2.95 & 1.21 & 1.23 & 2.81 & 1.33 & 1.67 & 0.00 & 2.78 & 2.45 & 2.64 & 2.04 \\
AFL (Transformer) & - & - & 1.97 & 1.68 & 0.86 & 0.00 & 1.38 & 1.84 & 1.27 & 1.61 & 1.47 & 1.82 & 1.01 & 1.28 & 1.35 \\
\arrayrulecolor{lightgray}\hdashline
\ourmethod{} & StarCoder-2    & 7B & 1.24 & 1.71 & 0.02 & 1.21 & 0.23 & 0.05 & 1.52 & 0.85 & 0.32 & 0.01 & 0.23 & 0.43 & 0.65\\
\ourmethod{} & StarCoder-2    &15B & 1.37 & 1.74 & 0.11 & 2.48 & 0.08 & 0.73 & 1.78 & 0.43 & 0.55 & 0.07 & 0.11 & 1.28 & 0.89 \\
\ourmethod{} & CodeLlama      & 7B & 1.62 & 1.32 & 0.18 & 1.15 & 0.49 & 2.43 & 0.75 & 0.19 & 0.37 & 0.05 & 0.14 & 1.15 & 0.82 \\
\ourmethod{} & DeepSeek-Coder & 7B & 1.74 & 1.10 & 0.50 & 2.00 & 1.21 & 3.45 & 6.84 & 1.70 & 3.45 & 1.44 & 1.63 & 1.47 & \cellcolor{red!25} \bf 2.21 \\
\ourmethod{} & CodeQwen       & 7B & 1.16 & 0.95 & 0.46 & 6.23 & 1.05 & 0.82 & 3.87 & 0.36 & 1.27 & 0.42 & 1.08 & 1.44 & 1.59 \\
\ourmethod{} & CodeShell      & 7B & 1.12 & 0.32 & 0.07 & 2.43 & 2.45 & 0.35 & 1.34 & 0.23 & 0.05 & 0.34 & 0.13 & 0.92 & 0.81 \\ \arrayrulecolor{black}\midrule
\rowcolor[rgb]{0.85,0.92,0.83}\multicolumn{16}{c}{\texttt{NM}} \\ \midrule
AFL (Original) & - & - & 1.35 & 0.41 & 0.04 & 0.38 & 2.03 & 1.29 & 0.10 & 0.01 & 0.00 & 0.23 & 0.03 & 0.05 & 0.49 \\
AFL (LSTM) & - & - & 1.95 & 0.84 & 0.09 & 9.74 & 0.00 & 0.90 & 2.47 & 0.00 & 0.00 & 0.24 & 0.75 & 0.72 & 1.47 \\
AFL (Transformer) & - & - & 0.90 & 0.83 & 0.30 & 3.48 & 1.27 & 1.31 & 3.80 & 1.32 & 0.00 & 0.00 & 1.29 & 0.52 & 1.25 \\
\arrayrulecolor{lightgray}\hdashline
\ourmethod{} & StarCoder-2    & 7B & 1.34 & 0.23 & 0.75 & 0.18 & 0.85 & 0.38 & 1.78 & 0.01 & 0.34 & 0.05 & 0.11 & 0.01 & 0.50 \\
\ourmethod{} & StarCoder-2    &15B & 1.41 & 0.37 & 1.21 & 0.34 & 0.93 & 0.72 & 2.43 & 0.08 & 0.17 & 0.14 & 0.05 & 0.05 & 0.66\\
\ourmethod{} & CodeLlama      & 7B & 0.17 & 0.13 & 0.83 & 0.71 & 0.71 & 0.81 & 1.82 & 0.03 & 0.11 & 0.35 & 0.08 & 0.26 & 0.50\\
\ourmethod{} & DeepSeek-Coder & 7B & 2.19 & 1.83 & 1.01 & 1.88 & 1.25 & 0.97 & 2.40 & 1.87 & 3.42 & 2.96 & 1.66 & 0.44 & \cellcolor{red!25} \bf 1.82 \\
\ourmethod{} & CodeQwen       & 7B & 1.83 & 0.54 & 1.27 & 1.39 & 1.37 & 1.32 & 2.98 & 0.97 & 2.41 & 1.12 & 2.69 & 2.43 & 1.69\\
\ourmethod{} & CodeShell      & 7B & 1.91 & 0.23 & 0.83 & 1.01 & 0.91 & 0.24 & 0.95 & 1.34 & 0.85 & 0.23 & 1.34 & 1.23 & 0.92
\\ \arrayrulecolor{black}\midrule
\rowcolor[rgb]{0.85,0.92,0.83}\multicolumn{16}{c}{\texttt{LINT\_XML}} \\ \midrule
AFL (Original) & - & - & 11.21 & 1.75 & 1.49 & 0.13 & 3.37 & 5.42 & 0.82 & 0.11 & 0.00 & 1.13 & 0.24 & 0.08 & 2.15 \\
AFL (LSTM) & - & - & 2.82 & 2.06 & 4.60 & 0.00 & 3.09 & 0.00 & 3.01 & 0.00 & 0.00 & 4.64 & 3.24 & 0.00 & 1.96 \\
AFL (Transformer) & - & - & 5.71 & 2.90 & 3.01 & 0.00 & 2.99 & 3.08 & 2.82 & 0.00 & 0.00 & 7.15 & 0.00 & 0.00 & 2.31 \\
\arrayrulecolor{lightgray}\hdashline
\ourmethod{} & StarCoder-2    & 7B  & 0.05 & 0.25 & 0.43 & 3.42 & 1.02 & 3.42 & 0.55 & 0.73 & 0.01 & 0.53 & 2.41 & 1.31 & 1.18 \\
\ourmethod{} & StarCoder-2    & 15B & 0.13 & 0.13 & 0.54 & 2.72 & 1.73 & 2.43 & 0.48 & 0.54 & 0.34 & 0.71 & 3.42 & 2.33 & 1.29 \\
\ourmethod{} & CodeLlama      & 7B  & 0.31 & 0.32 & 1.31 & 12.31 & 2.43 & 1.27 & 0.83 & 0.34 & 0.45 & 0.65 & 2.45 & 1.43 & 2.01 \\
\ourmethod{} & DeepSeek-Coder & 7B  & 0.99 & 0.00 & 0.49 & 14.28 & 8.31 & 0.36 & 0.84 & 0.72 & 0.41 & 2.61 & 1.42 & 9.80 & \cellcolor{red!25} \bf 3.35 \\
\ourmethod{} & CodeQwen       & 7B  & 0.68 & 0.82 & 0.19 & 19.51 & 6.42 & 0.00 & 1.65 & 0.91 & 0.28 & 3.63 & 0.41 & 2.51 & 3.08 \\
\ourmethod{} & CodeShell      & 7B  & 0.13 & 0.15 & 0.08 & 5.41 & 4.65 & 2.43 & 0.94 & 0.45 & 0.34 & 0.12 & 0.71 & 3.41 & 1.57  \\ \arrayrulecolor{black}\midrule
\rowcolor[rgb]{0.85,0.92,0.83}\multicolumn{16}{c}{\texttt{MP3\_GAIN}} \\ \midrule
AFL (Original) & - & - & 0.65 & 0.22 & 0.15 & 0.09 & 0.91 & 0.40 & 0.08 & 0.09 & 0.01 & 0.23 & 0.28 & 0.17 & 0.27 \\
AFL (LSTM) & - & - & 1.60 & 1.68 & 1.19 & 0.33 & 0.65 & 0.00 & 1.95 & 1.61 & 0.00 & 1.16 & 3.46 & 3.44 & 1.42 \\
AFL (Transformer) & - & - & 2.70 & 1.01 & 0.93 & 0.00 & 0.52 & 0.19 & 1.25 & 0.17 & 0.00 & 1.02 & 3.20 & 3.87 & 1.24 \\
\arrayrulecolor{lightgray}\hdashline
\ourmethod{} & StarCoder-2    & 7B  & 0.85 & 0.78 & 0.45 & 2.10 & 0.02 & 0.03 & 5.67 & 0.01 & 0.01 & 0.95 & 3.25 & 4.00 & 1.51 \\
\ourmethod{} & StarCoder-2    & 15B & 0.90 & 0.82 & 0.50 & 2.20 & 0.03 & 0.04 & 5.80 & 0.01 & 0.01 & 1.00 & 3.30 & 4.10 & 1.56 \\
\ourmethod{} & CodeLlama      & 7B  & 0.80 & 0.76 & 0.40 & 2.00 & 0.01 & 0.02 & 5.50 & 0.00 & 0.01 & 0.90 & 3.20 & 3.90 & 1.46 \\
\ourmethod{} & DeepSeek-Coder & 7B  & 0.76 & 0.75 & 0.36 & 2.13 & 0.00 & 0.00 & 6.44 & 0.00 & 0.00 & 1.25 & 3.30 & 4.12 & \cellcolor{red!25} \bf 1.59 \\
\ourmethod{} & CodeQwen       & 7B  & 1.09 & 0.83 & 0.48 & 0.82 & 1.05 & 0.00 & 2.72 & 0.00 & 0.00 & 1.72 & 3.21 & 3.50 & 1.29 \\
\ourmethod{} & CodeShell      & 7B  & 0.88 & 0.79 & 0.42 & 2.05 & 0.01 & 0.02 & 5.60 & 0.00 & 0.01 & 1.05 & 3.22 & 3.95 & 1.50 \\ \arrayrulecolor{black}\midrule
\rowcolor[rgb]{0.85,0.92,0.83}\multicolumn{16}{c}{\texttt{IMAGE\_MAGICK}} \\ \midrule
AFL (Original) & - & - & 1.95 & 0.30 & 0.36 & 1.89 & 1.14 & 2.26 & 0.74 & 0.00 & 0.09 & 0.94 & 0.16 & 0.09 & 0.83 \\
AFL (LSTM) & - & - & 3.12 & 1.29 & 0.26 & 0.00 & 0.00 & 0.00 & 5.66 & 0.00 & 0.00 & 0.00 & 0.00 & 13.39 & 1.98 \\
AFL (Transformer) & - & - & 3.88 & 1.05 & 0.62 & 3.02 & 1.67 & 1.22 & 12.28 & 0.00 & 0.00 & 2.34 & 1.16 & 0.00 & 2.27 \\
\arrayrulecolor{lightgray}\hdashline
\ourmethod{} & StarCoder-2    & 7B  & 2.05 & 1.82 & 0.70 & 1.40 & 0.00 & 0.80 & 8.90 & 1.30 & 0.00 & 3.20 & 8.10 & 3.15 & 2.62 \\
\ourmethod{} & StarCoder-2    & 15B & 2.25 & 2.00 & 0.75 & 1.50 & 0.00 & 0.85 & 9.05 & 1.40 & 0.00 & 3.30 & 8.20 & 3.25 & 2.71 \\
\ourmethod{} & CodeLlama      & 7B  & 2.10 & 1.85 & 0.71 & 1.42 & 0.00 & 0.82 & 8.92 & 1.32 & 0.00 & 3.22 & 8.12 & 3.17 & 2.64 \\
\ourmethod{} & DeepSeek-Coder & 7B  & 2.15 & 1.88 & 0.72 & 1.43 & 0.00 & 0.81 & 8.95 & 1.34 & 0.00 & 3.24 & 8.15 & 3.19 & 2.65 \\
\ourmethod{} & CodeQwen       & 7B  & 3.16 & 0.60 & 0.52 & 2.37 & 0.00 & 10.33 & 15.34 & 0.00 & 0.00 & 2.11 & 6.09 & 9.88 & \cellcolor{red!25} \bf 4.20 \\
\ourmethod{} & CodeShell      & 7B  & 2.12 & 1.86 & 0.73 & 1.44 & 0.00 & 0.83 & 8.97 & 1.35 & 0.00 & 3.25 & 8.16 & 3.20 & 2.66 \\\arrayrulecolor{black}\midrule
\rowcolor[rgb]{0.85,0.92,0.83}\multicolumn{16}{c}{\texttt{SPLIT\_TIFF}} \\ \midrule
AFL (Original) & - & - & 0.80 & 0.28 & 0.05 & 0.03 & 0.00 & 2.25 & 0.29 & 0.05 & 0.01 & 0.04 & 0.10 & 0.08 & 0.33 \\
AFL (LSTM) & - & - & 0.00 & 0.00 & 0.00 & 0.00 & 0.00 & 0.18 & 0.00 & 0.00 & 0.00 & 0.00 & 0.30 & 0.18 & 0.05 \\
AFL (Transformer) & - & - & 0.06 & 0.02 & 0.01 & 0.26 & 0.00 & 0.00 & 0.36 & 0.14 & 0.00 & 0.01 & 0.25 & 0.73 & 0.15 \\
\arrayrulecolor{lightgray}\hdashline
\ourmethod{} & StarCoder-2    & 7B  & 0.15 & 0.05 & 0.10 & 2.10 & 0.00 & 0.70 & 0.05 & 0.00 & 0.00 & 0.01 & 0.02 & 0.01 & 0.27 \\
\ourmethod{} & StarCoder-2    & 15B & 0.20 & 0.08 & 0.18 & 2.20 & 0.00 & 0.75 & 0.06 & 0.00 & 0.00 & 0.02 & 0.03 & 0.02 & 0.29 \\
\ourmethod{} & CodeLlama      & 7B  & 0.18 & 0.09 & 0.15 & 2.15 & 0.00 & 0.73 & 0.07 & 0.00 & 0.00 & 0.03 & 0.01 & 0.03 & 0.29 \\
\ourmethod{} & DeepSeek-Coder & 7B  & 0.34 & 1.01 & 0.22 & 2.33 & 0.43 & 0.76 & 0.04 & 1.08 & 0.44 & 0.54 & 0.64 & 0.34 & \cellcolor{red!25} \bf 0.68 \\
\ourmethod{} & CodeQwen       & 7B  & 0.23 & 0.10 & 0.00 & 0.00 & 0.00 & 0.00 & 0.19 & 0.00 & 0.00 & 0.00 & 0.26 & 0.19 & 0.08 \\
\ourmethod{} & CodeShell      & 7B  & 0.14 & 0.07 & 0.11 & 2.12 & 0.00 & 0.72 & 0.03 & 0.00 & 0.00 & 0.01 & 0.02 & 0.01 & 0.27 \\
 \arrayrulecolor{black}\midrule
\rowcolor[rgb]{0.85,0.92,0.83}\multicolumn{16}{c}{\texttt{TRAN\_JPEG}} \\ \midrule
AFL (Original) & - & - & 1.41 & 0.35 & 0.15 & 0.27 & 0.41 & 1.18 & 0.18 & 0.08 & 0.01 & 0.32 & 0.21 & 0.11 & 0.39 \\
AFL (LSTM) & - & - & 2.68 & 0.98 & 0.52 & 0.82 & 0.00 & 0.00 & 5.80 & 0.94 & 0.00 & 1.44 & 3.67 & 2.15 & 1.58 \\
AFL (Transformer) & - & - & 0.14 & 1.11 & 0.66 & 1.32 & 1.30 & 1.94 & 2.42 & 1.96 & 0.00 & 1.83 & 2.82 & 2.76 & 1.52 \\
\arrayrulecolor{lightgray}\hdashline
\ourmethod{} & StarCoder-2    & 7B  & 0.40 & 0.22 & 0.60 & 0.10 & 0.00 & 0.05 & 2.60 & 0.00 & 0.00 & 0.05 & 0.55 & 2.50 & 0.59 \\
\ourmethod{} & StarCoder-2    & 15B & 0.50 & 0.28 & 0.65 & 0.15 & 0.00 & 0.08 & 2.70 & 0.01 & 0.00 & 0.10 & 0.60 & 2.60 & 0.64 \\
\ourmethod{} & CodeLlama      & 7B  & 0.45 & 0.25 & 0.58 & 0.12 & 0.00 & 0.07 & 2.55 & 0.00 & 0.00 & 0.07 & 0.54 & 2.45 & 0.59 \\
\ourmethod{} & DeepSeek-Coder & 7B  & 0.36 & 0.21 & 0.56 & 0.00 & 0.00 & 0.00 & 2.52 & 0.00 & 0.00 & 0.00 & 0.53 & 2.40 & 0.55 \\
\ourmethod{} & CodeQwen       & 7B  & 3.40 & 0.54 & 0.86 & 0.45 & 0.53 & 0.54 & 1.29 & 1.13 & 0.54 & 2.11 & 6.21 & 1.34 & \cellcolor{red!25} \bf 1.58 \\
\ourmethod{} & CodeShell      & 7B  & 0.42 & 0.23 & 0.54 & 0.08 & 0.00 & 0.06 & 2.50 & 0.00 & 0.00 & 0.03 & 0.52 & 2.35 & 0.56 \\
\arrayrulecolor{black}\bottomrule
\end{tabular}
}
\caption{Evaluation results (EPM, \textperthousand{}) of multiple models. Bitflip $a/b$ denotes $a*b$ bits are flipped as a whole. Arith $a/b$ denotes the $a*b$ bits for addition and subtraction operations.}
\label{tab:EPM-results}
\vspace{-10pt}
\end{center}
\end{table*}

\begin{table*}[t!]
\centering
\resizebox{0.8\textwidth}{!}{
\begin{tabular}{llcccccccccc}
\toprule
Method & Base & Size & \texttt{READ\_ELF} & \texttt{OBJ\_DUMP} & \texttt{NM} & \texttt{LINT\_XML} & \texttt{MP3\_GAIN} & \texttt{IMAGE\_MAGICK} & \texttt{SPLIT\_TIFF} & \texttt{TRAN\_JPEG} & Avg.\\  \midrule
AFL (Original)              & -      & -  & 0  & 0  & 0   & 117 & 68  & 0  & 95  & 0  & 35 \\
AFL (LSTM)                  & -      & -  & 0  & 0  & 0   & 55  & 53  & 0  & 42  & 0  & 19 \\
AFL (Transformer)           & -      & -  & 0  & 0  & 0   & 61  & 45  & 0  & 77  & 0  & 23 \\
\ourmethod{} & StarCoder-2    & 7B          & 2  & 3  & 1  & 100 & 150 & 12 & 110  & 1  & 47 \\
\ourmethod{}  & StarCoder-2    & 15B         & \cellcolor{red!25} \bf\bf 4  & 5  & 2  & 120 & 180 & \cellcolor{red!25} \bf\bf 15 & 130  & 2  & 57 \\
\ourmethod{}  & CodeLlama      & 7B          & 3  & 2  & 0  & 90  & 140 & 10 & 100  & 1  & 43 \\
\ourmethod{}  & DeepSeek-Coder & 7B          & 2  & 4  & 0  & \cellcolor{red!25} \bf\bf 130 & \cellcolor{red!25} \bf\bf 230 & 3  & \cellcolor{red!25} \bf 224  & \cellcolor{red!25}\bf 3  & 75 \\
\ourmethod{}  & CodeQwen       & 7B          & 1  & \cellcolor{red!25} \bf\bf 9  & 0  & 114 & 209 & 4  & 221  & 2  & \cellcolor{red!25} \bf\bf 70 \\
\ourmethod{} & CodeShell      & 7B          & 3  & 6  & 1  & 95  & 160 & 11 & 105  & 1  & 48 \\
\bottomrule
\end{tabular}
}
\caption{Number of crashes of different models on eight datasets.}
\label{tab:NC-results}
\end{table*}

\begin{table*}[h!]
\centering
\resizebox{0.9\textwidth}{!}{
\begin{tabular}{ccccccccccccccccc}
\toprule
& \multicolumn{4}{c}{\texttt{READ\_ELF}} & \multicolumn{4}{c}{\texttt{OBJ\_DUMP}}     & \multicolumn{4}{c}{\texttt{NM}}          & \multicolumn{3}{c}{\texttt{LINT\_XML}} &      \\ \cmidrule{2-17}
& Line       & Branch       & Function       & Avg.       & Line        & Branch        & Function        & Avg.        & Line         & Branch       & Function       & Avg.       & Line            & Branch           & Function           & Avg. \\ \midrule
AFL (Original)  & 7.9 & 7.3 & 9.9 & 8.4 & 1.7 & 1.1 & 2.8 & 1.9 & 0.3 & 0.1 & 1.1 & 0.5 & 8.2 & 8.0 & 11.0 & 9.1   \\
AFL (LSTM)     & 7.3 & 6.6 & 9.0 & 7.6 & 1.6 & 1.0 & 2.8 & 1.8 & 0.3 & 0.2 & 1.1 & 0.5 & 8.1 & 7.8 & 10.9 & 8.9 \\
AFL (Transformer) & 6.6 & 5.9 & 8.2 & 6.9 & 1.6 & 1.0 & 2.7 & 1.8 & 0.3 & 0.1 & 1.0 & 0.5 & 8.0 & 7.7 & 11.0 & 8.9 \\
\ourmethod{} (Deepseek-Coder) & 14.9 & 16.5 & 15.4 & 15.6 & 2.0 & 1.5 & 3.1 & 2.2 & 0.6 & 0.3 & 1.9 & 0.9 & 9.2 & 9.4 & 11.8 & 10.1 \\
\ourmethod{} (CodeQwen)       & 14.5 & 15.9 & 15.2 & 15.2 & 2.0 & 1.5 & 3.1 & 2.2 & 0.6 & 0.4 & 1.9 & 1.0 & 8.7 & 8.8 & 11.3 & 9.6  \\ \midrule
                            & \multicolumn{4}{c}{\texttt{MP3\_GAIN}} & \multicolumn{4}{c}{\texttt{IMAGE\_MAGICK}} & \multicolumn{4}{c}{\texttt{SPLIT\_TIFF}} & \multicolumn{4}{c}{\texttt{TRAN\_JPEG}}       \\ \midrule
AFL (Original)              & 53.5 & 41.3 & 58.1 & 51.0 & 87.5 & 50.0 & 100.0 & 79.2 & 1.0 & 1.4 & 1.4 & 1.3 & 17.8 & 22.6 & 27.5 & 22.6 \\
AFL (LSTM)  &  53.2 & 40.8 & 58.1 & 50.7 & 87.5 & 50.0 & 100.0 & 79.2 & 0.9 & 1.3 & 1.1 & 1.1 & 15.5 & 18.8 & 26.3 & 20.2 \\
AFL (Transformer) &    54.0 & 41.5 & 58.1 & 51.2 & 87.5 & 50.0 & 100.0 & 79.2 & 1.0 & 1.4 & 1.4 & 1.3 & 15.4 & 18.3 & 26.3 & 20.0    \\
\ourmethod{} (Deepseek-Coder) & 54.9 & 43.2 & 59.1 & 52.4 & 87.5 & 50.0 & 100.0 & 79.2 & 1.0 & 1.6 & 1.4 & 1.3 & 19.0 & 24.7 & 27.9 & 23.9 \\
\ourmethod{} (CodeQwen) & 54.9 & 42.8 & 59.1 & 52.3 & 87.5 & 50.0 & 100.0 & 79.2 & 1.0 & 1.6 & 1.4 & 1.3 & 18.2 & 23.1 & 27.2 & 22.8   \\ \bottomrule
\end{tabular}}
\caption{Coverate rate (\%) of different models on 8 datasets.}
\vspace{-10pt}
\label{tab:LBC-results}
\end{table*}

\begin{figure*}[h!]
\begin{center} 
\includegraphics[width=0.7\textwidth]{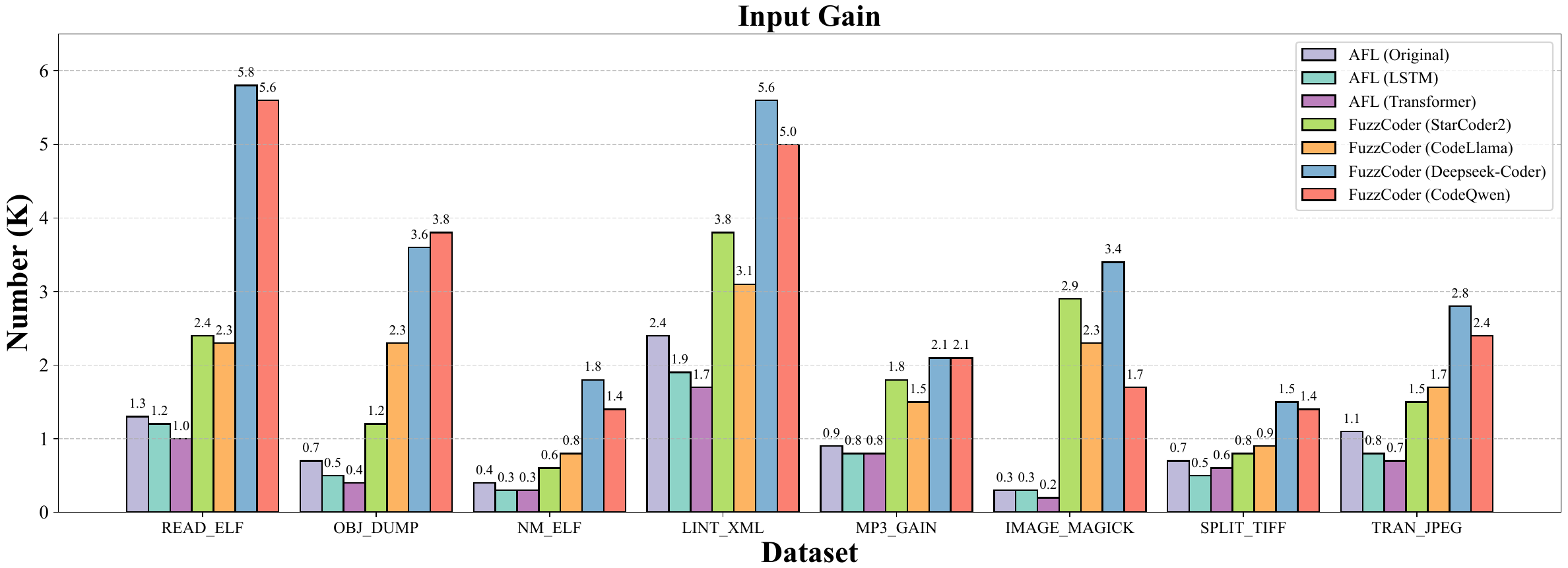}
\caption{Comparison between the baselines and \ourmethod{}.}
\label{fig:ig}
\vspace{-10pt}
\end{center}
\end{figure*}

\begin{figure*}[h!]
\centering
\includegraphics[width=0.6\textwidth]{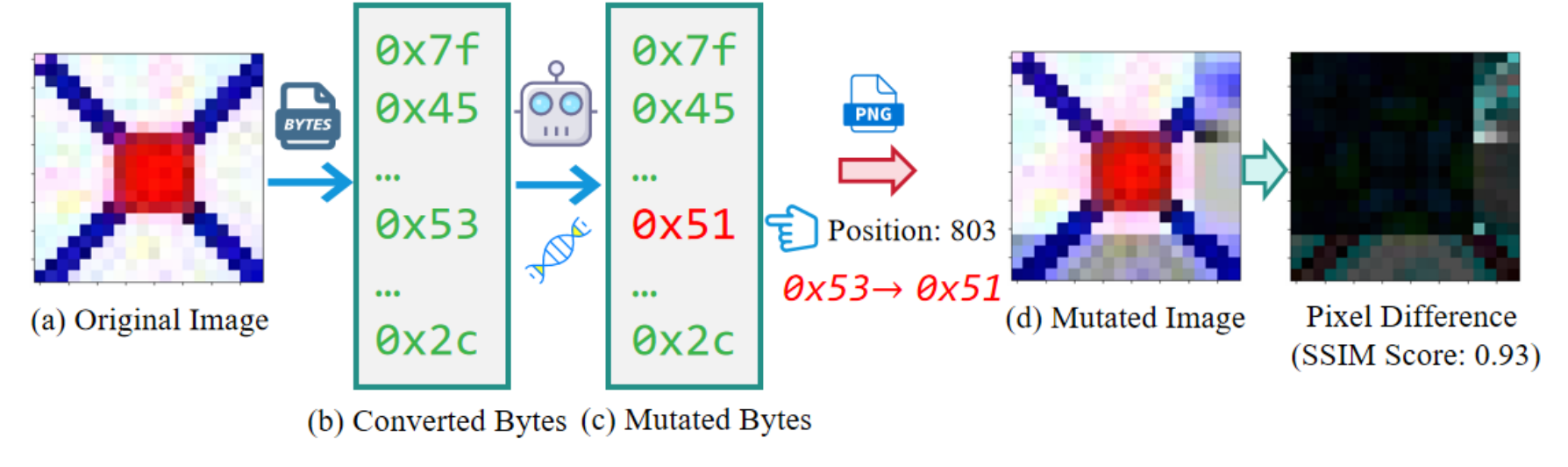}
	\caption{Comparison between the original JPG file and the JPG file after blur test}
	\label{fig:example}
\vspace{-10pt}
\end{figure*}

The AFL tool will first compile our test program and then use the test cases after mutation as input into the compiled program. The mutated test case causing a crash or triggering a new path will be used as seeds. \ourmethod{} adopts the Top-p sampling strategy to produce the candidate mutation strategy and position for diversity, which ensures that the effective mutation strategy and mutation positions are covered as much as possible.



\section{Experiments}
We evaluate our proposed method \ourmethod{} on 8 test sets, including \texttt{NM\_ELF}, \texttt{READ\_ELF}, \texttt{OBJDUMP\_ELF}, \texttt{LINT\_XML}, \texttt{MP3GAIN\_MP3}, \texttt{IMAGEMAGICK\_GIF}, \texttt{SPLIT\_TIFF}, and \texttt{TRAN\_JPEG}. In this section, we provide the details, results, and analysis of the experiments.

\subsection{Implementation Details}
By performing fuzzy tests using AFL\footnote{\url{https://lcamtuf.coredump.cx/afl/}}, we collect the original and variant inputs of successful attacks as a training set (nearly 30K SFT pairs). Our model based on open-source code LLMs CodeLlama, Deepseek-Coder, and CodeQwen is trained for 3 epochs with a cosine scheduler, starting at a learning rate of 5e-5 (3\% warmup steps). We use the AdamW~\citep{adamw} optimizer with a batch size of 1024 (max length 4K).

\subsection{Methods}
\textbf{AFL (Original):} The original AFL with the heuristic mutation rules is used as a baseline. 
\textbf{AFL (LSTM):} We use the encoder-decoder-based LSTM network without pre-training to decide the mutation position and strategy.
\textbf{AFL (Transformer):} The encoder-decoder-based Transformer without pre-training is incorporated into the AFL tool to improve the effectiveness of the fuzzing test.
\textbf{StarCoder-2:} StarCoder-2 models with 3B, 7B, and 15B parameters are trained on 3.3 to 4.3 trillion tokens, supporting hundreds of programming languages.
\textbf{Code-Llama:} Code-Llama is a family of code large language models based on Llama 2, providing infilling and long context capabilities.
\textbf{DeepSeek-Coder:} Deepseek-coder is a series of open-source code models with sizes from 1.3B to 33B, pre-trained from scratch on 2 trillion tokens.
\textbf{CodeQwen:} CodeQwen with 7B parameters supports 92 languages and 64K tokens.

\subsection{Evaluation Metrics}
\paragraph{Effective proportion of mutation (EPM):} 
For each mutation of the seed sample in the queue, a mutation location is selected, and then the corresponding mutation strategy is carried out for a mutation location. The effective proportion of mutations (\textperthousand) can be used to evaluate the effectiveness of different methods.

\paragraph{Number of Crashes (NC):}
This indicator refers to the number of input samples that cause the program to crash during fuzz testing and is used to measure the number of malicious inputs and the number of vulnerabilities.

\subsection{Main Results}

\paragraph{Results of EPM} In Table \ref{tab:EPM-results}, we find that the \ourmethod{} generally has better EPM than the AFL (Original) in each of the 8 programs and different LLMs have their own advantages in different programs. The results demonstrate that the code LLMs with the powerful understanding and generation capabilities can further bring improvement for the fuzzing test, compared to the AFL with small models.

\paragraph{Results of NC} 
In Table \ref{tab:NC-results}, our vulnerability findings for \texttt{READ\_ELF} and \texttt{NM} programs have 0 results on AFL (Original), AFL (LSTM and Transformer), which indicates that these two datasets are hard to vulnerabilities in the limited time. It shows that the mutation sequences from the LLMs easily lead to the crash for the program to be tested.

\section{Discussions and Analysis}
\paragraph{Input Gain (IG)}
Figure \ref{fig:ig} shows the number of new paths of changes in the execution of code blocks found during fuzz testing of the target program. We can observe that \ourmethod{} significantly improves the performance compared to the heuristic methods.

\paragraph{Coverage Rate}
In Table \ref{tab:LBC-results}, we report the coverage rate of different models, including line coverage, branch coverage, and function coverage. Line coverage refers to the ratio of whether each line of code has been executed at the time of the program under test fuzzing, and branch coverage refers to the ratio of whether each conditional branch has been executed at the time of the program under test fuzzing. By looking at these two metrics, we can know whether the test cases mutated by the Fuzzer can trigger more complete paths more effectively, so the higher these two metrics, the better.

\paragraph{Case study} In Figure \ref{fig:example}, we take the \texttt{JPEG\_TRANS} program as an example. In Figure \ref{fig:example}, the original Image will get Mutated Image after several rounds of fuzzing test. We use the big language model to guide the mutation of Image. For example, where Original Image was 0x53, it becomes 0x51. And the SSIM Score of Mutated Image vs. Original Image is 0.93. The Mutated Image is then fed into the JPEGTRAN program, which triggers a new code path or a program crash.

\section{Related Work}

\paragraph{Fuzzing Test}
Inspired by the success of sequence-to-sequence learning (s2s) in many NLP tasks \cite{transformer,alm,um4,hlt_mt}, the fuzzing test approaches use s2s to train neural networks to learn generative models of the input formats for fuzzing. For different input formats and the target program, random mutation of the inputs makes it hard to find the vulnerable positions to fuzz the program. Deep-learning-based methods \cite{learn_fuzz,learn_fuzz2,learn_fuzz3,yang2024seq2seq} present a technique to use LSTMs to learn grammar for PDF objects using a character-level model, which can then be sampled to generate new inputs. Instead of learning grammar, our technique uses neural networks to learn a function to predict promising locations in a seed file to perform mutations. The previous methods are hindered by a small number of parameters and the training corpora lack common knowledge of the byte sequence, codes, and reasoning. Recently, researchers \cite{fuzz4all,zero_shot_fuzzer} directly leverage prompt engineering to inspire the instruct-following capability of LLMs for effective fuzzing.

\paragraph{Domain-specific Large Language Model}
Large language models (LLMs) \cite{llama,llama2,gpt4,qwen} based on the decoder-only Transformer architecture have become a cornerstone in the realm of natural language processing (NLP). The pre-training on a vast corpus of internet text, encompassing billions of tokens enables LLMs to understand and generate human-style responses, making them highly versatile as zero-short learners. Further, code LLMs tailored for
software engineering tasks push boundaries of code understanding and generation \cite{chai2024xcot,guo2023owl,guo2024lemur,codellama,deepseek_coder,bGPT,slagle2024spacebyte}. The code LLM supports many code-related works, such as code translation, code generation, code refinement, program repair, and fuzzing. 
Recent methods tailored for fuzzing \cite{fuzz4all,yao2024fuzzllm,zero_shot_fuzzer} relying on common LLMs without domain-specific instruction tuning can not effectively unleash the potential of LLMs in the field of fuzzing.

\section{Conclusions}
In this paper, we present \ourmethod{}, a series of fine-tuned large language models for the fuzzing test. First, we collect the \instruct{} dataset based on a self-instruct strategy, which contains multiple programs to improve the generalization ability of LLMs on fuzzing operations. Then, to easily evaluate the performance of existing LLMs on fuzzing test, we also introduce the \benchmark{} evaluation benchmark dataset with eight programs. Besides, we
also introduce the mixture-of-adapter strategy to further enhance the instruction tuning
performance. Moreover, extensive experimental results on our \ourmethod{} demonstrate the
effectiveness of our \ourmethod{} for fuzzing test.

\clearpage
\bibliography{custom}
\bibliographystyle{acl_natbib}

\clearpage

\end{document}